\documentclass[letterpaper, 10 pt, conference]{ieeeconf}  

\IEEEoverridecommandlockouts                              

\usepackage{caption}
\usepackage{graphicx}
\usepackage[T1]{fontenc}
\usepackage[utf8]{inputenc}
\usepackage{microtype}
\usepackage{amsmath,amssymb,amsfonts,mathtools}
\usepackage{bm}
\usepackage[inkscapelatex=false]{svg}
\usepackage{cite}

\title{\LARGE \bf
FD-VLA: Force-Distilled Vision-Language-Action Model for Contact-Rich Manipulation
}

\author{Ruiteng Zhao$^{1}$, Wenshuo Wang$^{1}$, Yicheng Ma$^{2}$, Xiaocong Li$^{3}$, Francis E.H. Tay$^{4}$, \\Marcelo H. Ang Jr.$^{4}$ and Haiyue Zhu$^{5\dagger}$
\thanks{This research is supported by National Robotics Programme (NRP) 2.0 funding initiative "Domain-specific Robotics Foundation Models for Manufacturing (DS-RFM)".}
\thanks{$^{1}$Ruiteng Zhao and Wenshuo Wang are with the Advanced Robotics Centre, National University of Singapore, and are also attached students of SIMTech, A*STAR.
(email: \{ruiteng, wenshuo\_wang\}@u.nus.edu)}
\thanks{$^{2}$Yicheng Ma is with School of Electrical \& Electronic Engineering, Nanyang Technological University (NTU) and is also an attached student of SIMTech, A*STAR. 
(email: yichatma@gmail.com)}
\thanks{$^{3}$Xiaocong Li is with the College of Information Science and Technology, Eastern Institute of Technology, Ningbo, Ningbo 315200, China, and also with the John A. Paulson School of Engineering and Applied Sciences, Harvard University, Cambridge, MA 02134 USA. (email: xiaocongli@eitech.edu.cn, xiaocongli@seas.harvard.edu).}
\thanks{$^{4}$Francis E.H. Tay and Marcelo H. Ang Jr are with Advanced Robotics Centre at National University of Singapore, Singapore 117608,
{(email: mpetayeh@nus.edu.sg, mpeangh@nus.edu.sg)}}%
\thanks{$^{5}$Haiyue Zhu is with the Singapore Institute of Manufacturing Technology, Agency for Science, Technology and Research (A*STAR), Singapore 138634,
(email: zhu\_haiyue@simtech.a-star.edu.sg)}
\thanks{$\dagger$ Corresponding author: zhu\_haiyue@simtech.a-star.edu.sg}
}

\begin{document}

\maketitle
\thispagestyle{empty}
\pagestyle{empty}

\begin{abstract}


Force sensing is a crucial modality for Vision-Language-Action (VLA) frameworks, as it enables fine-grained perception and dexterous manipulation in contact-rich tasks. We present Force-Distilled VLA (FD-VLA), a novel framework that integrates force awareness into contact-rich manipulation without relying on physical force sensors. The core of our approach is a Force Distillation Module (FDM), which distills force by mapping a learnable query token, conditioned on visual observations and robot states, into a predicted force token aligned with the latent representation of actual force signals. During inference, this distilled force token is injected into the pretrained VLM, enabling force-aware reasoning while preserving the integrity of its vision-language semantics. This design provides two key benefits: first, it allows practical deployment across a wide range of robots that lack expensive or fragile force-torque sensors, thereby reducing hardware cost and complexity; second, the FDM introduces an additional force-vision-state fusion prior to the VLM, which improves cross-modal alignment and enhances perception-action robustness in contact-rich scenarios. Surprisingly, our physical experiments show that the distilled force token outperforms direct sensor force measurements as well as other baselines, which highlights the effectiveness of this force-distilled VLA approach.

\end{abstract}

\section{INTRODUCTION}

Vision-Language-Action (VLA) architectures~\cite{brohan2022rt, zitkovich2023rt, team2024octo, kim2024openvla, zhang2025spatial} have emerged as a dominant paradigm in robot imitation learning, enabling scalable manipulation skills for a wide spectrum of everyday and industrial tasks. Building on large pretrained Vision-Language Models (VLMs)~\cite{steiner2024paligemma, wang2024qwen2}, modern VLAs integrate perception, reasoning, planning, and control into a unified end-to-end framework. This allows them to map RGB inputs and natural-language instructions directly to low-level robot commands, while benefiting from strong semantic priors and broad generalization capabilities. Yet, real-world deployments reveal fundamental limits of vision-only policies: occlusions, lighting shifts, depth ambiguities, and subtle contact effects often remain poorly captured in images. This underscores the need for truly multimodal VLAs that combine additional sensory signals with visual and linguistic context.

\begin{figure*}[t]
    \centering
    \includegraphics[width=0.98\textwidth]{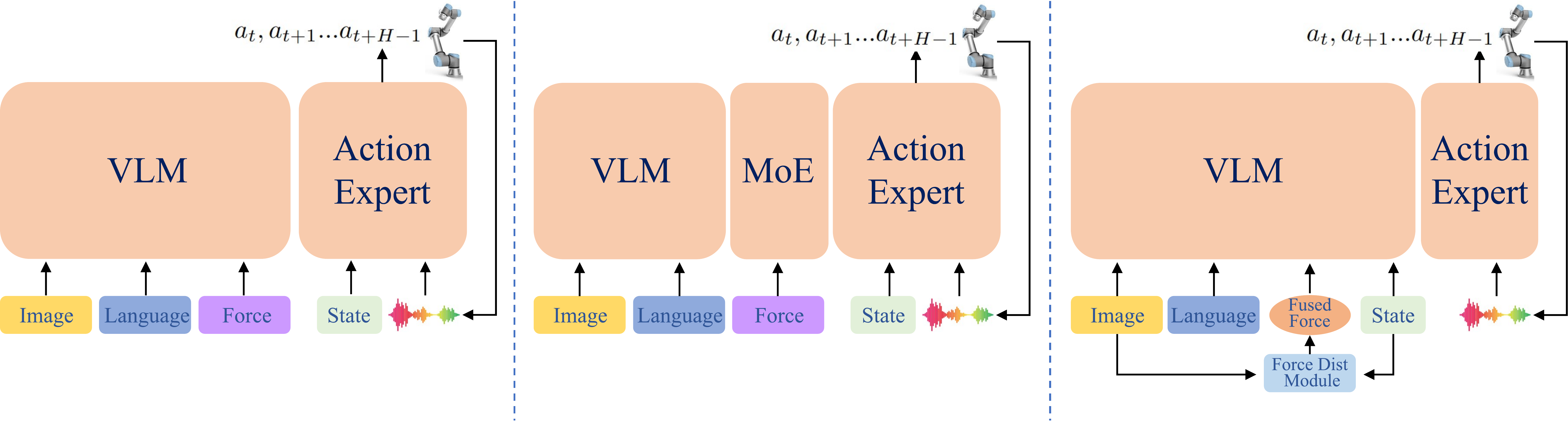}
    \caption{Overview of differentiate architectures of force VLAs. (\textbf{Left}) Tactile-VLA with tactile encoder directly encode tactile information. (\textbf{Middle}) Force-VLA with MoE module between VLM and action expert. (\textbf{Right}) Our FD-VLA using predicted force instead of actual force to get better fused feature and sensor free at deployment.}
    \label{fig:comparison}
\end{figure*}

Among the possible modalities, force/tactile sensing stands out as a particularly critical but still underexplored dimension for VLAs, as it provides direct information about contact dynamics, compliance, and physical interactions that are often invisible to vision~\cite{xie2025towards, funk2025importance}. Recent studies~\cite{yu2025forcevla, huang2025tactile, zhang2025ta, bi2025vla} have begun to incorporate force/tactile signals into VLAs, which raises a key question of how to efficiently incorporate this additional modality within the VLA framework. Fig.~\ref{fig:comparison} overviews the different architecture of those  VLAs with force/tactile modality. Tactile-VLA encodes raw force signals into compact latent embeddings that are projected and fused with pretrained vision-language features via cross-modal alignment layers, which enriches visuomotor reasoning with contact awareness. However, it risks disturbing the pretrained VLM’s carefully aligned vision-language semantics through modality misalignment and catastrophic interference. Instead, ForceVLA encodes force signals into a dedicated token that is injected post-VLM and fused through a force-aware Mixture-of-Experts (MoE) module, thereby enriching the policy with contact-sensitive cues while its late fusion preserves the integrity of the pretrained vision-language representations. Though effective, its MoE design potentially increases the risk of training instability as well as inference complexity. Moreover, the late-fusion mechanism limits fine-grained force-vision-state interactions, reducing tight perception-action coupling and generalization.

In this work, we propose a novel FD-VLA framework that incorporates a distilled force token, rather than raw sensor signals, into the VLA model to enhance contact-rich manipulation. The distillation is achieved via a Force Distillation Module (FDM) that maps a learnable query token, conditioned on visual observations and robot states through attention, into a latent predicted force token. During the training, the supervision of FDM is achieved by aligning the predicted force token with the latent representation of actual force signals. At inference time, this distilled force token can then be injected into the VLM, enabling force-aware reasoning without requiring direct sensor measurements. This enables practical deployment on a wide range of robot platforms that lack force sensors, reducing hardware cost and complexity while still benefiting from force-aware reasoning. More important, FDM provides an additional force-vision-state fusion in the force distillation before VLM, which improves the consistency of cross-modal alignment and enhances the perception-action robustness in contact-rich manipulation. 

In summary, the main contributions of this work are summarized as follows: 
\begin{itemize}
    \item We propose the FD-VLA framework that injects a distilled force token into the VLA model to improve contact-rich manipulation.
    \item The design of FDM is introduced for VLA to distill predicted force token by conditioning learnable query token on vision and state inputs.
    \item Our approach enables force-aware reasoning without requiring physical sensors, while provides additional force-vision-state cross-modal alignment.
\end{itemize}

\section{RELATED WORKS}

\subsection{Vision-Language-Action Models}

Compared to traditional robot grasp learning~\cite{wang2024sgsin, asif2018graspnet, liu2024relationgrasp}, The emergence of large-scale VLA models has redefined the landscape of robot learning by coupling pretrained vision-language representations with action generation. Pioneering systems~\cite{brohan2022rt, zitkovich2023rt, cheang2024gr, black2024pi_0, team2024octo, kim2024openvla, wen2025tinyvla, shukor2025smolvla, liu2025hybridvla} built on large-scale datasets~\cite{o2024open, khazatsky2024droid, walke2023bridgedata} demonstrated the potential of aligning visual-linguistic embeddings with robotic policies, which achieves strong generalization across diverse tasks and platforms. These models typically rely on large pretrained backbones, enabling them to parse natural language instructions and interpret complex visual scenes. Recent studies have moved beyond the conventional vision-language setting, introducing additional modalities such as audio information~\cite{zhao2025vlas} and depth information~\cite{li2025pointvla, zhen20243d} to provide richer context for robotic perception and action generation. However, for contact-rich manipulation tasks, these additional modalities are less effective than force or tactile sensing, which provide direct measurements of physical interactions. 


\subsection{Force Integration in Contact-Rich Manipulation}

Contact-rich manipulation tasks require precise modeling of physical interactions such as contact forces, slips, and deformations, which are difficult to infer from visual observations alone. Traditional vision-only approaches often struggle with unstable dynamics and lack the fine-grained feedback necessary for reliable control~\cite{chi2023diffusion, wang2024graspcontrast}. To address these challenges, recent works have begun to integrate force and tactile sensing into robotic policies~\cite{wu2025tacdiffusion, liu2025forcemimic, he2025foar, hou2025adaptive, xue2025reactive}, demonstrating significant improvements in stability and precision. VLAs inherit the same fundamental limitation as traditional vision-centric policies: they lack direct access to physical interaction cues. In order to solve VLA's limitation, recent research has begun to incorporate tactile and force sensing into the VLA paradigm. Examples include FuSe, ForceVLA, and Tactile-VLA~\cite{jones24fuse, yu2025forcevla, huang2025tactile}. These models explore auxiliary loss designs, modality-specific routing to capture embodied dynamics more faithfully or using prior knowledge and tactile sensors to achieve zero-shot generalization. Unlike current works that extend VLAs through heavy finetuning, architectural modifications or relying on special sensors, we explore enhancing model's ability in contact-rich tasks by using images and robot state to generate force information and sensor-free at deployment.

\begin{figure*}[t]
    \centering
    \includegraphics[width=0.98\textwidth]{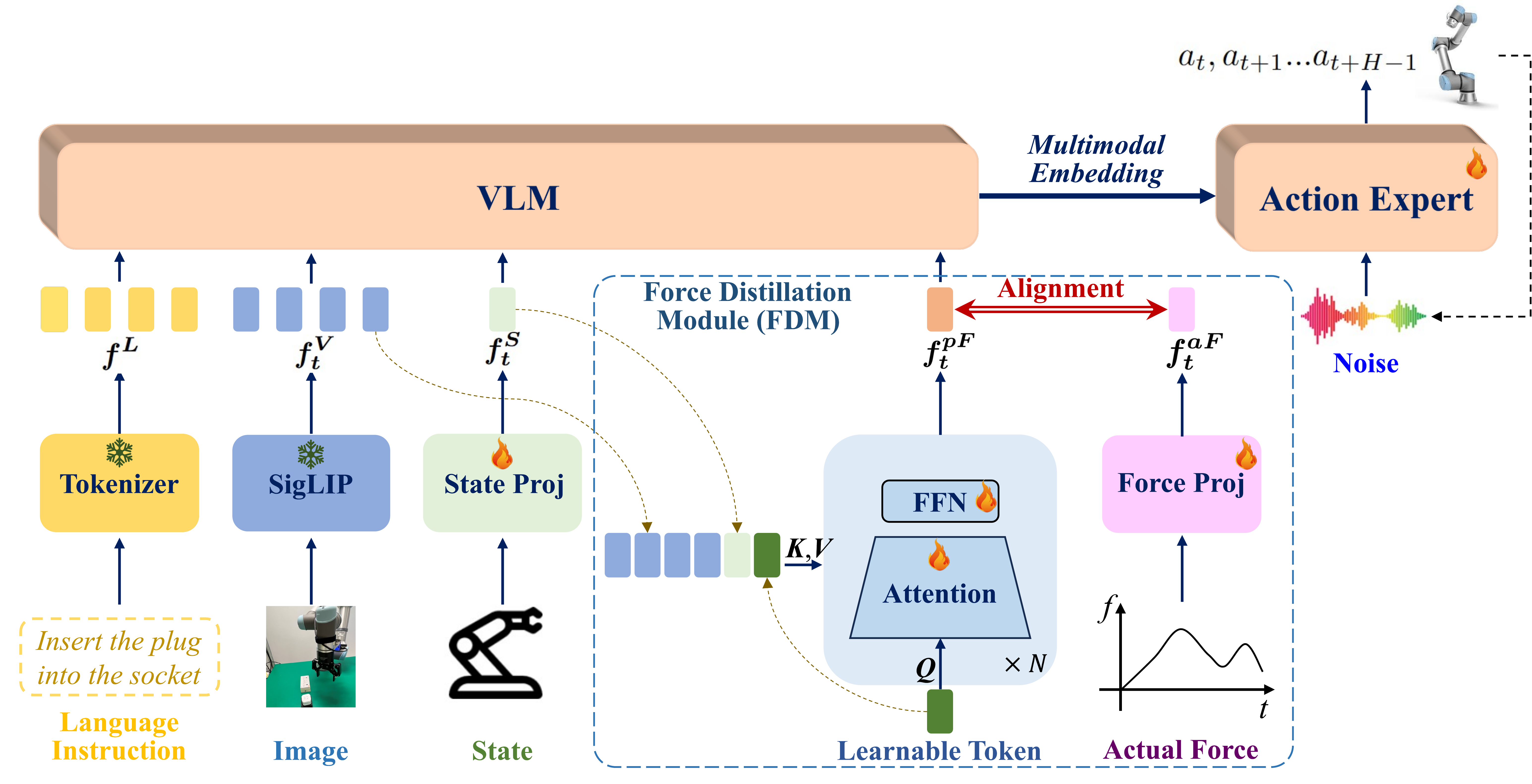}
    \caption{Overview of our framework. During training, measured force signals are encoded into an actual force token via a lightweight projection. A learnable query attends to image and state tokens to predict a latent force token, which is supervised against the actual force token. At inference, the model synthesizes this latent force representations solely from vision and state inputs, eliminating the need for tactile hardware. The predicted force token is then fused with language, image, and state tokens inside the pretrained VLM, and an action expert consumes the fused representation to generate action sequences.}
    \label{fig:overview}
\end{figure*}

\subsection{Cross-Modal Knowledge Distillation}
Transferring information across modalities is essential for bridging semantic perception and physical interaction in robotic systems. Cross-modal knowledge distillation provides a principled way to align heterogeneous sensory streams, allowing one modality to benefit from the complementary strengths of another. Earlier studies~\cite{hinton2015distilling, gupta2016cross} explored feature-level transfer through carefully designed objectives, laying the foundation for cross-modal knowledge distillation. Subsequent works~\cite{wang2018learning} injected depth priors to enhance the visual representation for 3D perception tasks. And~\cite{xia2023robust} introduced an adaptive transfer approach to achieve more robust and effective cross-modal knowledge distillation.
Currently, little attention has been given to contact-rich robotic manipulation, where force or tactile sensing remains expensive and sparsely available.
Our work tackles this challenge by distilling implicit force representations from vision and state inputs, eliminating the need for dedicated tactile hardware at deployment.




\section{METHODOLOGY}

This section presents the FD-VLA framework with detail methodologies. 

\subsection{Motivation}

The objective of this work is to explore an effective and data-efficient approach to incorporate additional force modality into VLA model for contact-rich manipulation. Directly retraining large vision-language models to accommodate an additional modality is prohibitively expensive and risks degrading the well-aligned visual-linguistic representations. We seek a design that integrates force without modifying the pretrained VLM, thereby preserving its semantic alignment while avoiding costly retraining. However, the challenge is how to achieve good cross-modal fusion and alignment in the meantime. 

\begin{figure}[!b]
    \centering
    \includegraphics[width=0.48\textwidth]{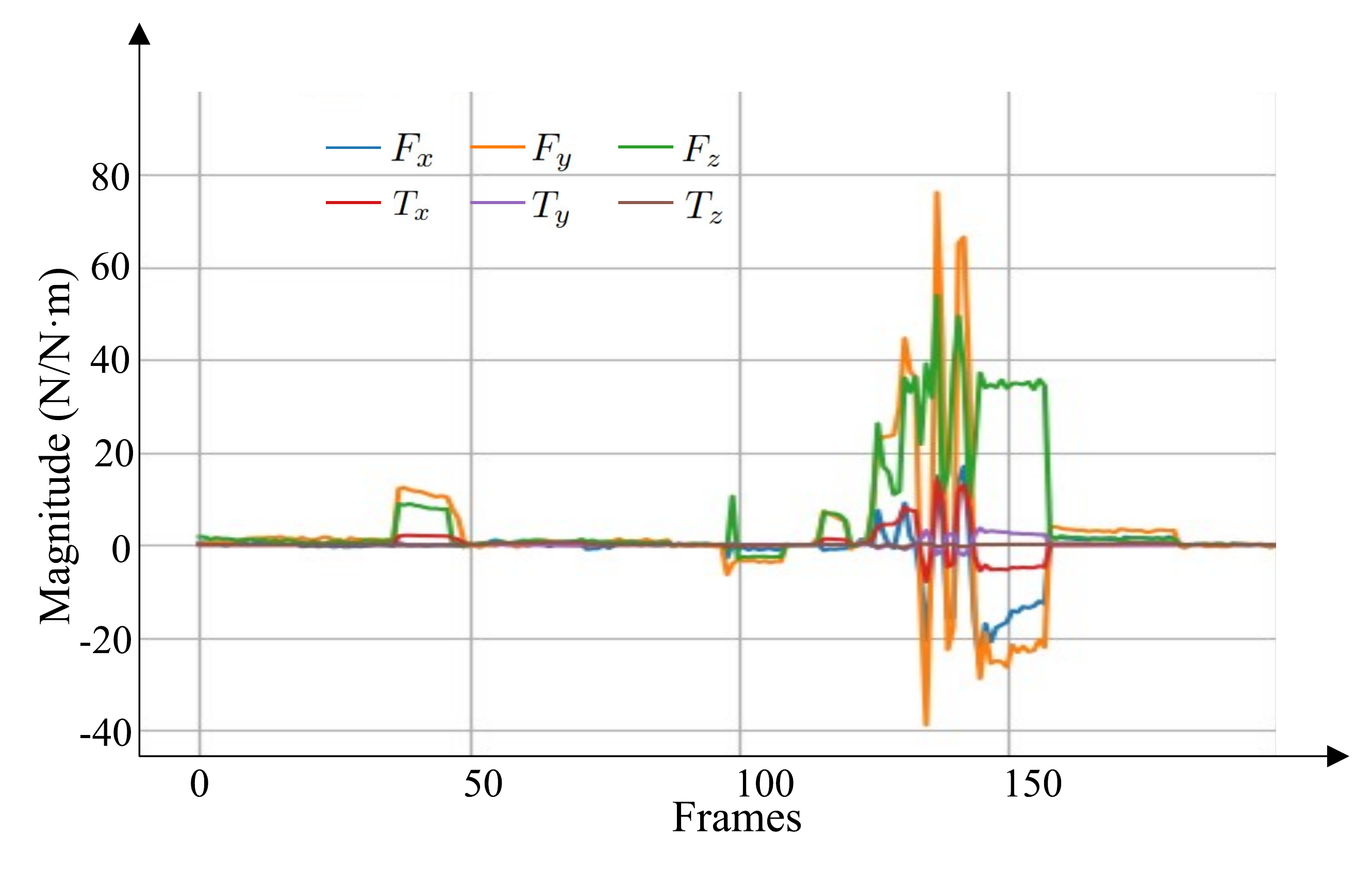}
    \caption{Visualization of raw force in the plug insertion task.}
    \label{fig:raw_force}
\end{figure}

Moreover, we also observed that raw force signals collected in real-world settings, as shown in Fig.~\ref{fig:raw_force}, may not be optimal for policy learning, as they are often corrupted by high-frequency artifacts and low-frequency drift. Motivated by these challenges, we propose FD-VLA guided by three design principles: (i) leverage predicted force tokens obtained through distillation with actual force signals during training only; (ii) maximize feature-level cross-modal fusion and alignment before, within, and after the VLM so as to preserve pretrained semantic priors while enriching action representations; and (iii) enable sensor-free inference at deployment, ensuring practical applicability across diverse robotic platforms.

\subsection{Overall Framework}
Fig.~\ref{fig:overview} illustrates the overall framework of the proposed FD-VLA. The multimodal inputs of VLA include language instruction $\bm{L}$, visual observation $\bm{V}_t$, robot state $\bm{S}_t$, and force $\bm{F}_t$, where $t$ denotes the timestamp. Following SmolVLA, we employ a pretrained vision-language model SmolVLM-2 with SignLIP~\cite{zhai2023sigmoid} as the perception backbone, and computations in the VLM are selectively skipped. First, the visual observation, language instruction, and robot proprioceptive state are encoded into their corresponding feature representations $\bm{f}_t^{V}\in\mathbb{R}^{N_v\times D}$, $\bm{f}^{L}\in\mathbb{R}^{N_l\times D}$, and $\bm{f}_t^{S}\in\mathbb{R}^{N_s\times D}$, respectively, by leveraging on the VLM’s pretrained tokenizer and vision encoder, and a lightweight projection layer consisting of a single-layer MLP. These embeddings are projected to match the hidden dimension $D$ of the VLM, ensuring compatibility for subsequent fusion and processing.

Rather than directly incorporating the raw force measurements into the VLM, we introduce the Force Distillation Module (FDM) that can predict a latent force representation from the real-time visual observation and robot proprioceptive state,
\begin{equation}
\bm{f}_t^{pF}=\mathrm{FDM}\big(\bm{f}_t^{V}, \bm{f}_t^{S}, \bm{p}\big),
\label{equ:attention_2}
\end{equation}
where $\bm{p}\in\mathbb{R}^{1\times D}$ is a learnable token for force prediction. Note that the $FDM$ is trained to align with real force signals $\bm{f}^{aF}\in\mathbb{R}^{1\times D}$ encoded by a projection layer only during the training stage, while at inference time it operates entirely without force measurements.

Next, all four modalities of inputs are concatenated along the token dimension, and fuse as
$\bm{f}_t^{fs}=[\bm{f}_t^{V},\bm{f}^{L},\bm{f}_t^{S},\bm{f}_t^{pF}]$ to pass into the VLM for multimodal reasoning. Within the VLM, the integrity of pretrained visual-linguistic knowledge are carefully preserved as the VLM parameters are frozen, and an attention masking strategy is employed to ensure that the state and predicted force tokens ($\bm{f}_t^{S}$ and $\bm{f}_t^{pF}$) remain disentangled from the core visual and language streams. This separation prevents interference with the pretrained semantic representations while still enabling the action head to integrate all modalities effectively for downstream control. Finally, an action expert follows a transformer-based policy head with a conditional flow-matching decoder that maps the fused multimodal embeddings to action sequences,
\begin{equation}
[a_t, a_{t+1},...,a_{t+H-1}]=\pi_{\theta}\big(\mathrm{VLM}(\bm{f}_t^{fs})\big),
\label{equ:action}
\end{equation}
where $\pi_{\theta}(\cdot)$ represents the action expert policy head. This architecture allows our system to leverage the semantic richness of pretrained VLM while introducing stable, task-relevant physical reasoning through force distillation, achieving both robustness and efficiency without additional sensing requirements.







\subsection{Force Distillation Module (FDM)}

Our FDM generates a compact, state-aware force representation that can be seamlessly integrated into the VLA pipeline without requiring specialized tactile hardware in inference. The key insight is that force and tactile information are not isolated measurements but are inherently coupled with the robot’s proprioceptive state and its visual manipulation context: variations in joint torques, velocities, and positions, together with visual cues such as object deformation and spatial changes, jointly reflect the underlying contact forces. By exploiting this strong correlation, FDM distills force representations directly from multimodal inputs that are already available on most robotic platforms, thereby eliminating the need for additional physical sensors. Furthermore, modeling this inherent mapping enhances cross-modal fusion and alignment and facilitate more accurate perception-action coupling in contact-rich tasks. Finally, FDM mitigates the noise and instability of raw sensor signals by learning a supervised latent embedding that serves as a denoised, task-relevant proxy for physical force, filtering out high-frequency artifacts while preserving the dynamics most critical for manipulation.

The realization of FDM consists of two parallel branches, i.e., the prediction branch and actual force branch. In the actual force branch, the raw force measurements collected from the robot’s force/torque sensor are encoded to $\bm{f}_t^{aF}$ by a lightweight MLP to project them into the same embedding space as the predicted force representation. This projected embedding serves as the supervisory signal for the prediction branch only during the training stage. 

In the prediction branch, the attention mechanism is utilized to predict the learnable token $\bm{p}\in\mathbb{R}^{1\times D}$ into the latent force feature, given the current images embeddings $\bm{f}_t^{V}$ and robot proprioceptive state embeddings $\bm{f}_t^{S}$. In detail, the design of prediction branch casts the force latent generation as a retrieval problem conditioned on the vision-state context by leveraging a single learned query $\bm{p}$. 
A context matrix is defined as 
$\bm{C}_t=[\,\bm{p},\bm{f}_t^{V},\bm{f}_t^{S}\,]\in\mathbb{R}^{N_{c}\times D}$, where $N_{c}=1+N_v+N_s$. 
Using a single-query multi-head attention, $\bm{p}$ serves as the only query $\bm{Q}=\bm{p}\bm{W}_Q\in\mathbb{R}^{H\times d_k}$ while $\bm{C}_t$ provides keys $\bm{K}=\bm{C}_t\bm{W}_K\in\mathbb{R}^{N\times H\times d_k}$ and values $\bm{V}=\bm{C}_t\bm{W}_V\in\mathbb{R}^{N\times H\times d_k}$. The output of the multi-head attention $\bm{Z}$ can be calculated as
\begin{equation}
\begin{split}
\bm{\alpha}_t=&\mathrm{softmax}\!\Big(\frac{(\bm{pW}_Q)([\,\bm{p},\bm{f}_t^{V},\bm{f}_t^{S\,}]\bm{W}_K)^\top}{\sqrt{d_k}}\Big),\\
    \bm{Z}=&\bm{\alpha}_t \bm{V}\in\mathbb{R}^{H\times d_k},
\end{split}
\label{equ:attention_4}
\end{equation}
where $\alpha_t$ represents a normalized weight vector indicating the relative importance of each token in the context. The final predicted force token can be evaluated as 
\begin{equation}
\bm{f}_t^{pF}=\mathrm{FFN}\!\big(\mathrm{LN}([\bm{Z}_1\!,\!\cdots\!,\!\bm{Z}_H]\bm{W_O}+\bm{p})\big).
\label{equ:attention_6}
\end{equation}
Including the query token $\bm{p}$ in both the key and value sets enables self-conditioning through the residual path, injecting the learned contact prior directly into the aggregation. 

Finally, the force distillation is achieved by the alignment between the feature representations from these two branches, where an auxiliary distillation loss is used as
\begin{equation}
\mathcal{L}_{\text{dist}}
= \big\| \bm{f}_t^{pF} - \bm{f}_t^{aF} \big\|_2^2.
\label{equ:attention_5}
\end{equation}

\begin{figure*}[t]
    \centering
    \includegraphics[width=0.98\textwidth]{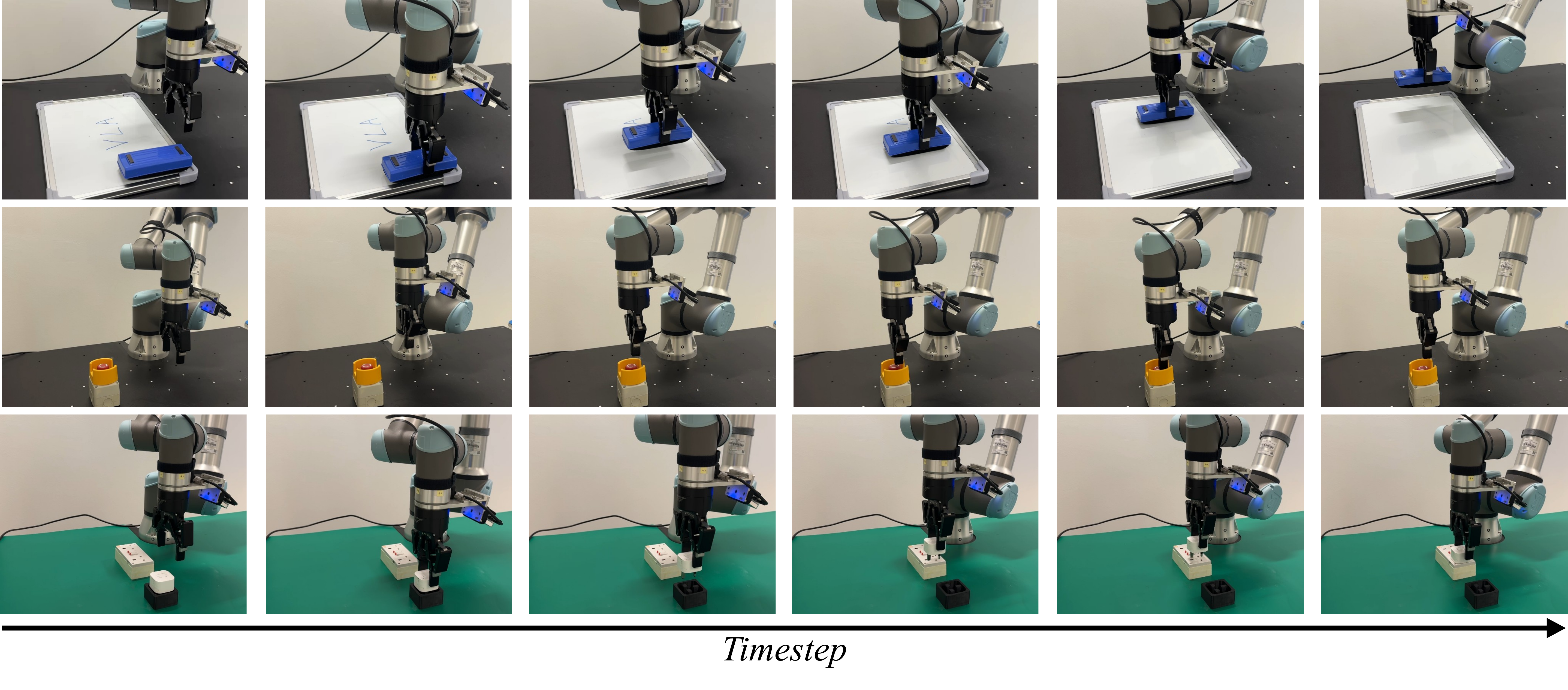}
    \caption{Visualization of real-world experimental tasks: 1) Clean the whiteboard, 2) Press the emergency button, 3) Insert the plug into the socket.}
    \label{fig:task}
\end{figure*}

\subsection{Directional Attention Masking Mechanism}


To incorporate state and force tokens into the pretrained VLM without corrupting its semantic priors, inspired by SmolVLA~\cite{shukor2025smolvla}, we adopt a directional attention masking mechanism. The key idea is to treat vision and language tokens as the frozen perceptual stream and state/force tokens as the control stream. 

The input four-modality tokens are divided into two streams, $\bm{X} = \{\bm{X}^{\text{percept}}, \bm{X}^{\text{control}}\}$, where 
$\bm{X}^{\text{percept}} = \{\bm{f}^V, \bm{f}^L\}$ are the frozen perceptual stream, and 
$\bm{X}^{\text{control}} = \{\bm{f}^S, \bm{f}^{pF}\}$ are the control stream. To preserve pretrained semantics while enabling fusion, we construct a Directional Attention Mask $\bm{M} \in \{0,1\}^{N_c \times N_c}$, where
\begin{equation}
\bm{M}_{ij} =
\begin{cases}
1, & \text{if } x_i \in \bm{X}^{\text{percept}} \ \text{and } x_j \in \bm{X}^{\text{percept}}, \\
1, & \text{if } x_i \in \bm{X}^{\text{control}} \ \text{and } x_j \in \bm{X}^{\text{percept}}, \\
1, & \text{if } x_i \in \mathbf{X}^{\text{control}},\; x_j \in \mathbf{X}^{\text{control}},\; \text{and } i \geq j,
\\
0, & \text{otherwise}.
\end{cases}
\end{equation}

Applying the mask inside the VLM’s transformer yields masked self-attention, $\bm{X}'=\mathrm{SelfAttention}(\bm{X}, \bm{M})$. Specifically,
\begin{equation}
\begin{split}
    \bm{X}'^{\text{percept}}&=\mathrm{SelfAttention}(\bm{X}^{\text{percept}}),\\
    \bm{X}'^{\text{control}}&=\mathrm{SelfAttention}(\bm{X}^{\text{control}}, \bm{X}^{\text{percept}}).
\end{split}
\end{equation}
This design enforces one-way information flow, i.e., perceptual tokens can only attend to each other, preserving pretrained vision-language alignment, while control tokens can attend to perceptual tokens, thereby fusing proprioception and 
force with perceptual context for downstream action prediction.




\subsection{Action Expert}
The action expert $\mathbf{\pi}_\theta$ is instantiated as a transformer that predicts an action chunk
$\mathbf{A}_t=[a_t,...,a_{t+H-1}]$ conditioned on VLM features $\mathbf{X}_{t}$. We train the action expert with a conditional flow-matching objective,
\begin{equation}\label{eq:fm_loss}
\mathcal{L}^{\tau}(\theta)
= \mathbb{E}_{\,p(\mathbf{A}_t \mid \mathbf{X}_t),\; q(\mathbf{A}_t^{\tau} \mid \mathbf{A}_t)}
\Big[ \big\lVert \mathbf{v}_{\theta}(\mathbf{A}_t^{\tau}, \mathbf{X}_t)
      - \mathbf{u}(\mathbf{A}_t^{\tau} \mid \mathbf{A}_t) \big\rVert^{2} \Big],
\end{equation}
where $\mathbf{A}_\tau \;=\; \tau\,\mathbf{A} + (1-\tau)\,\epsilon$, with $\epsilon\!\sim\!\mathcal{N}(0,I)$. $\pi_\theta(\cdot)$ is trained to output the vector field $\mathbf{u}(\mathbf{A}_t^{\tau} \mid \mathbf{A}_t)=\epsilon-\mathbf{A}_t$ from the VLM features
and the noisy actions $\mathbf{A}_t^{\tau}$, $\tau$ is sampled from a Beta distribution. The policy learns a velocity field rather than stepwise residuals, which is well suited to chunked action prediction. 


\subsection{Overall Training Objective}
The overall objective of our FD-VLA framework combines two complementary components, i.e., a standard policy learning loss and a force-distillation loss, which is defined as
\begin{equation}
\mathcal{L}=\mathcal{L}^{\tau}(\theta)+\lambda \mathcal{L}_{\text{dist}},
\end{equation}
where $\lambda$ is the weight for controlling force supervision.

\section{EXPERIMENTS}

\begin{figure}[t]
    \centering
    \includegraphics[width=0.48\textwidth]{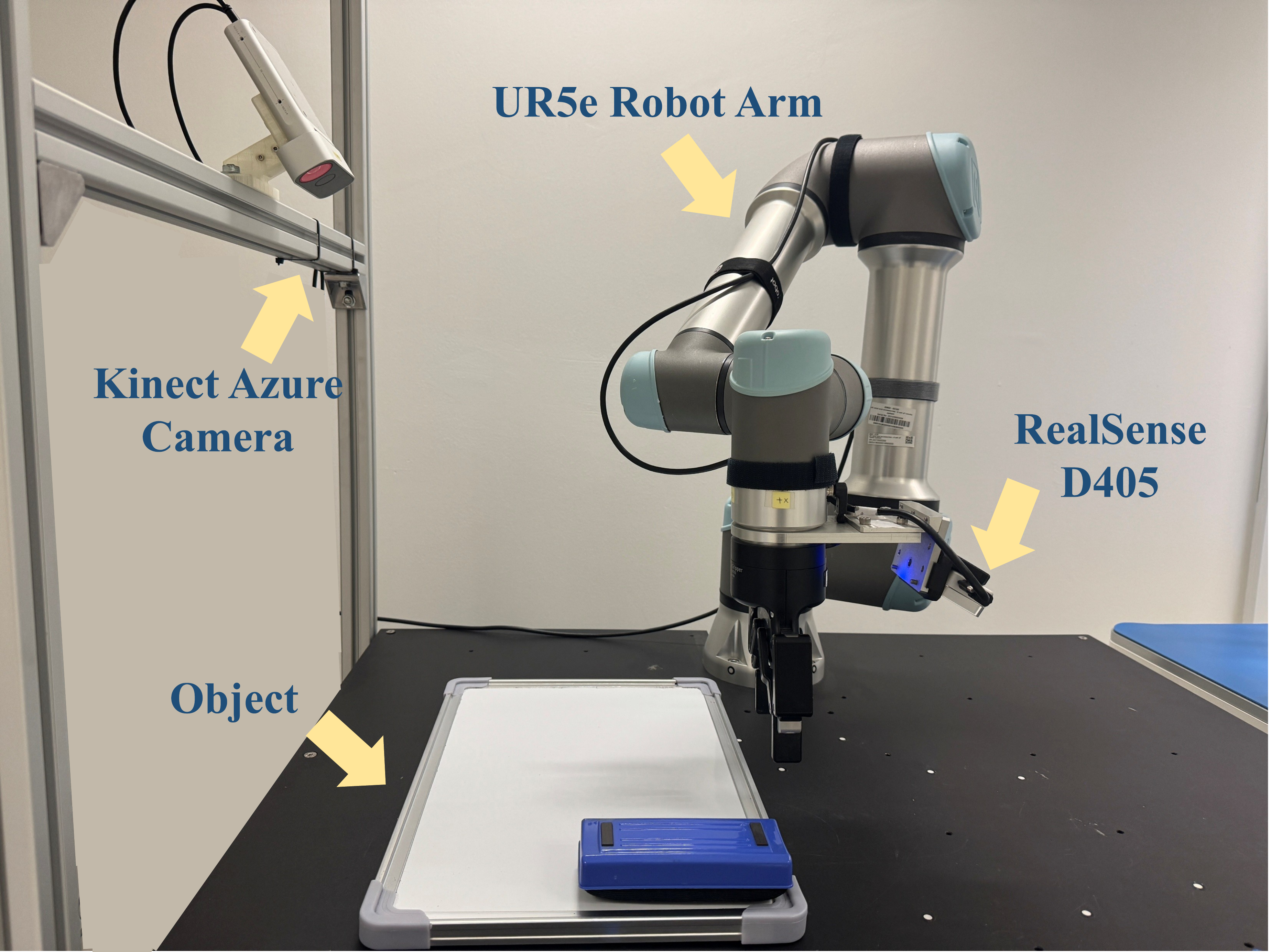}
    \caption{Visualization of the real robotic platform. We use a UR5e robot arm as the main manipulation platform, the Kinect Azure camera as the main camera and the RealSense D405 as the gripper camera.}
    \label{fig:real_setup}
\end{figure}

\begin{figure*}[t]
    \centering
    \includegraphics[width=0.98\textwidth]{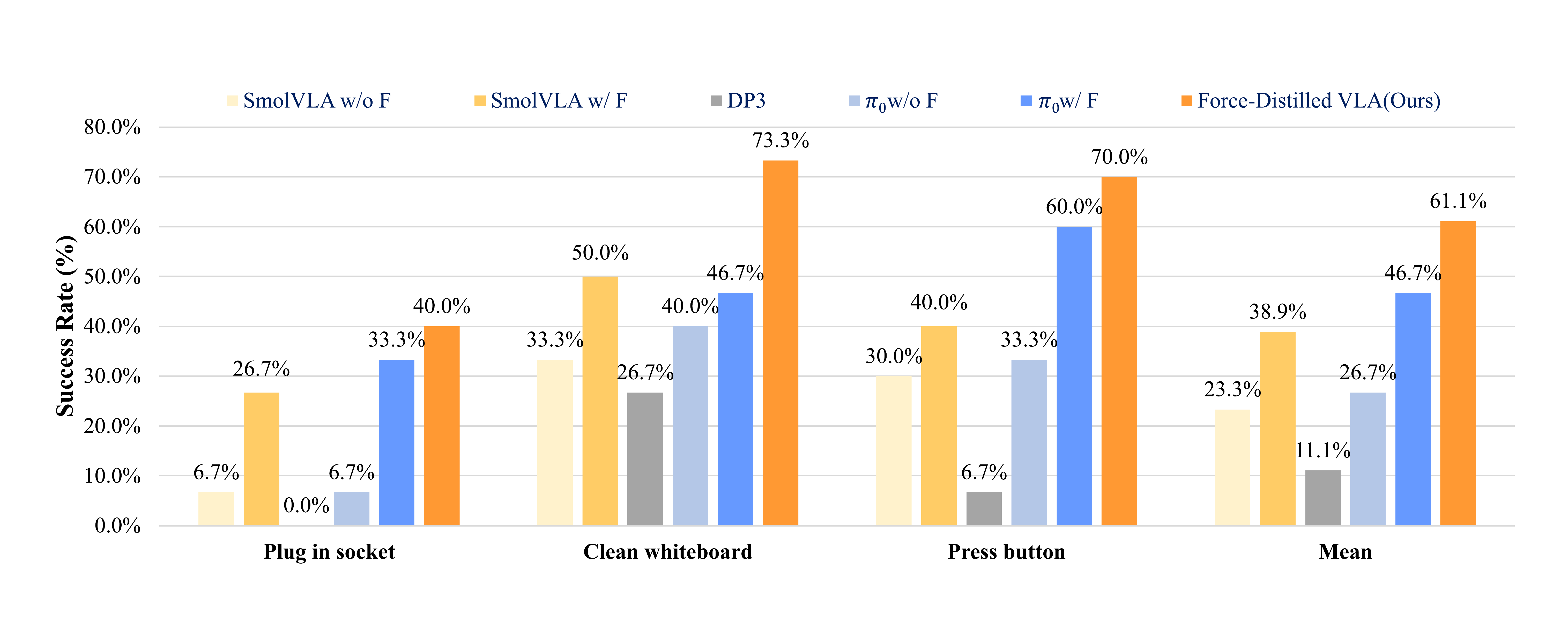}
    \caption{Success rates for three contact-rich manipulation tasks: Plug in Socket, Clean Whiteboard, and Press Button. Results are averaged over 30 evaluation episodes per task. We compare FD-VLA (ours) with SmolVLA, $\pi_0$ and DP3, SmolVLA and $\pi_0$ are evaluated with and without force inputs. Our model achieves consistently higher performance, which highlights the benefit of force distillation for accurate and robust manipulation.}
    \label{fig:results}
\end{figure*}

This section presents an extensive evaluation of the FD-VLA model through real-world experiments and analytical studies. 

\subsection{Experimental Setup}
In the experiments, we use a UR5e robotic arm as our main manipulation platform, an Azure Kinect camera for main RGB image acquisition and a RealSense D405 for robot gripper RGB image acquisition. The real experimental platform is shown in Fig.~\ref{fig:real_setup}. This platform provides a large workspace, high repeatability, and easy programmability. During operation, the robotic arm and the cameras work together through real-time data exchange to achieve precise  manipulation.

\subsection{Experimental  Tasks}

To evaluate the effectiveness of FD-VLA, we conducted experiments on three diverse contact-rich manipulation tasks: 1) Clean Whiteboard, demanding consistent motion planning with continuous contact across the surface. 2) Press Emergency Button, requiring precise vertical pressing and exceeding the resistance and spring position of the button. 3) Insert Plug, which requires precise positional alignment and regulated force application during engagement. The visualization of the tasks is shown in Fig.~\ref{fig:task}. During data collection, we use the force sensor inside the UR5e robot arm to obtain actual force information. To efficiently collect expert demonstrations, we use the 3Dconnexion SpaceMouse to teleoperate the robotic arm. We collect 50 demonstrations for each task. The demonstration procedure is standardized to ensure that all operators follow a predefined protocol consistently.

For evaluation, each task was trained using a set of 50 demonstrations and subsequently evaluated over 30 independent test trials to ensure statistical robustness. Success criteria were defined based on precise task-specific outcomes. For the Clean Whiteboard task, a trial was considered successful only if all visible markings on the whiteboard were completely erased. In the Press Emergency Button task, success required the button to be fully depressed and remain engaged without rebound. For the Insert Plug task, success was achieved when the plug was fully seated within the socket, indicating proper alignment and insertion.

\subsection{Main Results}

To ensure a fair and comprehensive evaluation, we benchmark FD-VLA against three representative policies: DP3~\cite{Ze2024DP3}, $\pi_0$~\cite{black2024pi_0}, and SmolVLA~\cite{shukor2025smolvla}. DP3 is selected as a strong diffusion-based control framework with a parameter scale comparable to ours, which provides a capacity-matched baseline that excels at generating precise motion trajectories. $\pi_0$ represents the state-of-the-art VLA paradigm with substantially larger capacity and powerful language-conditioned control. Including $\pi_0$ in baselines allows us to compare against a substantially larger model  with scale and broad pretraining. SmolVLA is a recent compact VLA model that integrates multimodal reasoning and control through a compact transformer-based design, which also serves as our foundational model. For all baselines, we standardize training data, evaluation tasks, and optimization budgets to control for dataset and compute confounds, and we report all evaluation metrics following identical experimental protocols to guarantee a rigorous and fair comparison between methods.


Fig.~\ref{fig:results} presents the success rates across three contact-rich manipulation tasks. For example, in Clean Whiteboard task our FD-VLA achieves a success rate of 73.3\%, which is at least 23.3\% higher than any other baselines. Across all the tasks, our FD-VLA achieves the highest overall performance with a mean success rate of 61.1\%, substantially outperforming both SmolVLA without force encoding (23.3\%), DP3 (11.1\%) and even $\pi_0$ without force encoding (46.7\%) which contains approximately ten times as many parameters as our model. 

To further evaluate the contribution of our force distillation approach, we compare FD-VLA with models trained using direct raw force inputs. By injecting raw force information, SmolVLA achieves an improvement of 15.6\% but still falls short of our method by 22.2\%. $\pi_0$ also shows an improvement of 20.0\%, yet our approach continues to outperform it by 14.4\% despite $\pi_0$ operating at a substantially larger parameter scale. These results suggest that simply providing raw force signals is insufficient for optimal performance, as the model must directly process noisy, high-dimensional sensory inputs. 
In contrast, our predicted fused force module integrates information from both vision and robot state, enabling the extraction of compact and task-relevant force representations. 
In addition, models trained with raw force require access to real-time force measurements during inference, which limits their applicability to platforms with specialized hardware.

Our approach overcomes this limitation by predicting force representations directly from visual and state inputs, eliminating the need for force sensors at test time while still achieving superior performance. 
This highlights both the effectiveness and practicality of our method for deploying contact-rich manipulation policies in real-world settings.

\subsection{Ablation Study}

To validate the proposed Force Distillation module, we conduct an ablation by replacing the learnable force tokens in the Force Distillation module with direct ground-truth force measurements obtained from sensors. We compare three variants on three contact-rich tasks, each with 30 independent rollouts under the same initialization, time budget, and success criteria as the main experiments: 1) Without FDM, directly using an MLP layer to encode force. 2) FDM with force token, replacing the learnable token with the actual force token encoded by an MLP layer. 3) FDM with learnable token, our proposed method.

Table~\ref{tab:efficiency} shows the clear improvement in adding FDM, which indicates that deep cross-modal interactions in the early fusion are a key driver for the improvements. From the table, we could also find that when replacing the learnable token with the latent force features encoded from ground-truth force measurements, the mean success rate drops. This result highlights the challenge of directly incorporating raw force: raw signals are often high-dimensional and noisy, which makes them difficult for the policy to integrate effectively. In contrast, the learnable token not only avoids the pitfalls of directly feeding fixed force features but also retains the benefits of real physical supervision, providing reliable contact information.

\begin{table}[t]
\renewcommand\arraystretch{1.4}
    \centering
    \caption{Ablation study of Force Distillation Module.}
    \resizebox{0.48\textwidth}{!}{
    \begin{tabular}{ccccc}
    \hline
    Model & Plug & Clean & Press & Mean \\
    \hline w/o FDM & $8/30$ & $15/30$ & $12/30$ & $38.9\%$ \\
    \hline FDM w/ force token & $12/30$ & $17/30$ & $17/30$ & $51.1\%$ \\    
    \hline FDM w/ learnable token & $12/30$ & $22/30$ & $21/30$ & $61.1\%$ \\    
    \hline
    \end{tabular}}
    \label{tab:efficiency}
\end{table}

\begin{figure}[t]
    \centering
    \includegraphics[width=0.48\textwidth]{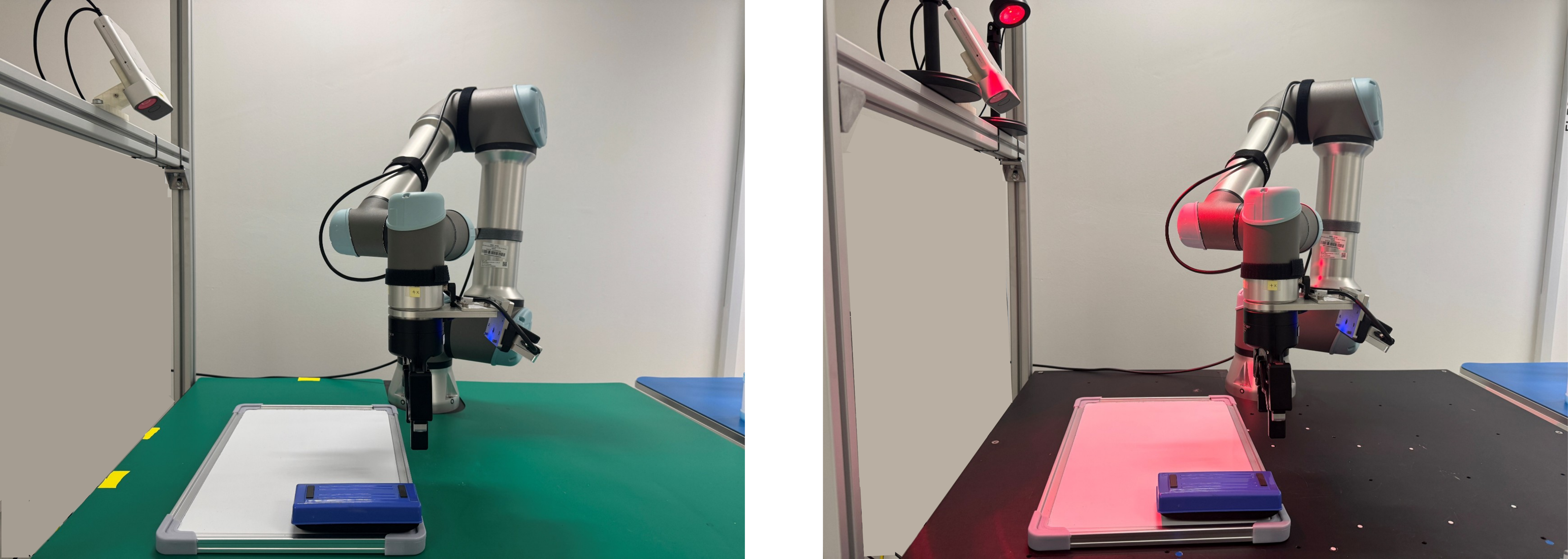}
    \caption{Illustration of our visual generalization settings: (\textbf{Left}) Novel Background, where the scene's background differs from training; and (\textbf{Right}) Visual Perturbation, which involves changes in colorful lights.}
    \label{fig:generlization}
\end{figure}

\subsection{Model Generalization}
Visual generalization denotes a robot’s ability to sustain task performance under distribution shifts in scene appearance. We study this challenge in two representative scenarios: (1) \textbf{Novel Background}, where layouts and textures differ from training, and (2) \textbf{Visual Perturbations}, where lighting, contrast, and color distributions are altered. As illustrated in Fig.~\ref{fig:generlization}, these settings reflect common real-world challenges, requiring robots to operate beyond the narrow visual domain of training data. Qualitative results show that our method remains robust across both scenarios. On novel backgrounds, the robot leverages structural cues rather than overfitting to textures. Under photometric perturbations, it adapts smoothly to illumination and appearance changes without significant policy degradation. These findings highlight the model’s capacity to extract task-relevant features that remain invariant to superficial visual shifts.

\section{CONCLUSIONS}
In this work, we introduce FD-VLA, a novel VLA architecture that integrates force awareness and is tailored for contact-rich manipulation tasks. Instead of directly relying on noisy and hardware-dependent raw force signals, our method employs a Force Distillation Module (FDM), which distills task-relevant and denoised force representations from vision and state inputs. We leverage attention masks to keep state and predicted force tokens separate from the pretrained VLM stream. This design prevents interference with semantic representations while achieving efficient multimodal integration with minimal additional parameters. The fused features are then consumed by the action expert to generate precise, contact-aware control. Experiments across diverse real-world tasks show that FD-VLA consistently outperforms baselines without force reasoning or with raw force inputs. We hope this work paves the way for advancing multimodal learning in VLAs toward complex manipulation tasks. Looking ahead, we plan to extend this framework to multi-robot coordination and move toward scalable, general-purpose manipulation systems.

\bibliographystyle{IEEEtran}
\bibliography{reference_copy}

\end{document}